\documentclass[runningheads]{llncs}
\usepackage[T1]{fontenc}
\usepackage{graphicx,verbatim}

\usepackage{epstopdf}
\usepackage{amssymb}
\usepackage{amsmath}
\usepackage{marvosym}
\usepackage{multirow}
\usepackage{hyperref}
\usepackage{color}
 \usepackage{booktabs}
 \usepackage{colortbl}

\usepackage{xcolor}

\begin{document}
\title{MAUP: Training-free Multi-center Adaptive Uncertainty-aware Prompting for Cross-domain Few-shot Medical Image Segmentation}
%
\begin{comment}  %% Removed for anonymized MICCAI 2025 submission
\author{First Author\inst{1}\orcidID{0000-1111-2222-3333} \and
Second Author\inst{2,3}\orcidID{1111-2222-3333-4444} \and
Third Author\inst{3}\orcidID{2222--3333-4444-5555}}
%
\authorrunning{F. Author et al.}
\institute{Princeton University, Princeton NJ 08544, USA \and
Springer Heidelberg, Tiergartenstr. 17, 69121 Heidelberg, Germany
\email{lncs@springer.com}\\
\url{http://www.springer.com/gp/computer-science/lncs} \and
ABC Institute, Rupert-Karls-University Heidelberg, Heidelberg, Germany\\
\email{\{abc,lncs\}@uni-heidelberg.de}}

\end{comment}

\author{Yazhou Zhu\inst{1} \and  %index{Zhu, Yazhou} 
Haofeng Zhang\inst{1} %index{Zhang, Haofeng}
\textsuperscript{(\Letter)}} 
\authorrunning{Y. Zhu et al.}

\institute{School of Computer Science and Engineering, Nanjing University of Science and Technology, Nanjing 210094, China. 
\email{\{zyz\_nj,zhanghf\}@njust.edu.cn}}

\maketitle              % typeset the header of the contribution
\begin{abstract}

Cross-domain Few-shot Medical Image Segmentation (CD-FSMIS) is a potential solution for segmenting medical images with limited annotation using knowledge from other domains. The significant performance of current CD-FSMIS models relies on the heavily training procedure over other source medical domains, which degrades the universality and ease of model deployment. With the development of large visual models of natural images, we propose a training-free CD-FSMIS model that introduces the Multi-center Adaptive Uncertainty-aware Prompting (MAUP) strategy for adapting the foundation model Segment Anything Model (SAM), which is trained with natural images, into the CD-FSMIS task. To be specific, MAUP consists of three key innovations: (1) K-means clustering based multi-center prompts generation for comprehensive spatial coverage, (2) uncertainty-aware prompts selection that focuses on the challenging regions, and (3) adaptive prompt optimization that can dynamically adjust according to the target region complexity. With the pre-trained DINOv2 feature encoder, MAUP achieves precise segmentation results across three medical datasets without any additional training compared with several conventional CD-FSMIS models and training-free FSMIS model. The source code is available at: \url{https://github.com/YazhouZhu19/MAUP}.


\end{abstract}

\section{Introduction}

Medical Image Segmentation (MIS) has become a foundational technique in modern healthcare, disease diagnosis and treatment planning \cite{azad2024review,zhu2024selfreg}. Traditional deep learning based MIS models fundamentally relay on the extensive annotated data which is both costly and time-consuming to acquire in medical imaging, especially in rare disease scans \cite{ZHANG2024102996,shaker2024unetr}. Besides, this problem is also compounded by the heterogeneity of medical scans, encompassing diverse modalities (e.g., MRI, CT and X-Ray) and anatomical structures, which leads to the hard of intra-class generalization in semantic segmentation \cite{jiang2024m4oe,you2024mine}. 

Accordingly, Few-shot Medical Image Segmentation (FSMIS) is proposed to address these limitations, enhancing the model capability of segmenting with limited annotated images \cite{zhu2024partition,tang2025few,zhang2024prototype}. 
%However, existing FSMIS models are mostly based on the prototypical network, introducing substantial training cost and limiting their adaptation ability in clinical deployment.
However, FSMIS still requires a large number of labeled samples in the same domain for model training \cite{cheng2025cascaded,song2024mtpnet}, even if the target class for testing differs, e.g., both training and testing must be performed on CT images. To solve this problem, some researchers have tried to implement model training using labeled samples from other medical domains and transfer them to the target medical domain \cite{zhu2025robustemddomainrobustmatching,yang2025cdsg}, e.g., training on CT and applying the model to MR images, which is called Cross-domain FSMIS (CD-FSMIS). Despite the effectiveness of these CD-FSMIS methods \cite{bo2025cross,Chen_2024_WACV,tang2024cross}, there are still two problems: (1) They require a large amount of precisely annotated images in the medical domains, which are also usually scarce and need to be annotated by professional doctors; (2) A lot of model training for each application is still necessary.

Recently, the Segment Anything Model (SAM) \cite{kirillov2023segment,huang2025sam} has created the possibilities for training-free segmentation model with appropriate prompting strategy \cite{wang2025osam}, yet its direct application to CD-FSMIS still faces several critical challenges: (1) Prompts (e.g., point prompts) need to capture the complete structure of complex medical objects; (2) Lack of boundary awareness leads to the segmentation ambiguity, especially in some low-contrast medical images, and (3) Traditional fixed prompting strategy always fails to adapt to varying anatomical complexities. In order to address these limitations, we introduce a novel training-free few-shot medical image segmentation model with \textbf{M}ulti-center \textbf{A}daptive \textbf{U}ncertainty-aware \textbf{P}rompting (\textbf{MAUP}) strategy for the SAM model. To be specific, the MAUP strategy employs the point prompts and it consists of four components: (1) A multi-center prompting strategy that employs K-means clustering on high-similarity regions to make sure the spatial diversity in point prompts, leading to the comprehensive coverage of complex anatomical structures; (2) Leveraging the morphology based periphery similarity maps to generate negative prompts, enhancing boundary delineation accuracy in low-contrast regions; (3) An uncertainty-aware prompt selection approach which identifies and focuses on challenging regions with analyzing variance value across similarity maps; and (4) An adaptive point prompts quantity determination method that dynamically adjusts the number of point prompts based on target complexity. 

Besides, MAUP employs the DINOv2 \cite{oquab2023dinov2} as the feature encoder in few-shot segmentation algorithm, enabling the both of discriminative and generalizable capability of core features without requirement of model training. And the synergy of proposed four components make MAUP effectively handle diverse medical imaging scenarios while maintaining the high segmentation accuracy. Our method achieves the precise segmentation results in three medical image datasets including the abdominal datasets \textbf{Abd-MRI} \cite{kavur2021chaos} and \textbf{Abd-CT} \cite{ABD-CT}, and the cardiac dataset \textbf{Card-MRI} \cite{zhuang2018multivariate}. In conclusion, the key contributions of our model can be summarized as follows:

(1) We introduce a training-free model which addresses the limitations of existing cross-domain few-shot medical image segmentation approaches via designed innovative prompting strategy.  

(2) We propose a prompting strategy MAUP which considers the spatial diversity, boundary awareness, uncertainty guidance and adaptive point prompts quantity adjustment for enhancing the segmentation accuracy.

(3) The proposed model achieves state-of-the-art performance on three medical imaging datasets widely used in research.

\begin{figure*}[!t]                   % htbp
\centering
\includegraphics[width=0.99\textwidth]{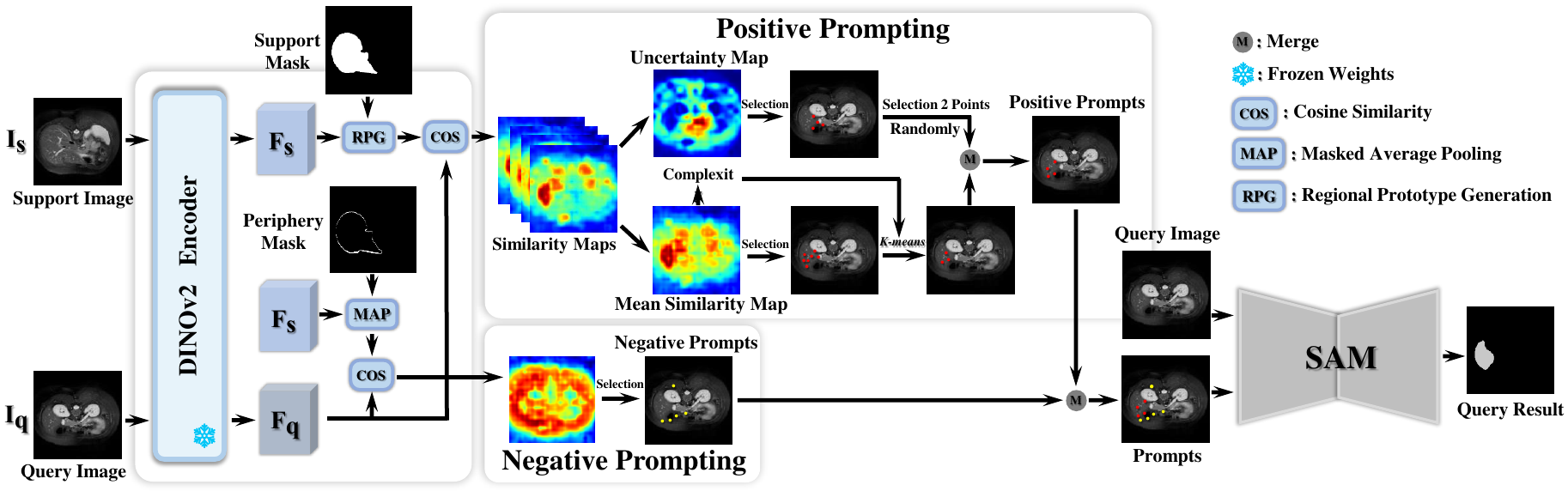}
\vspace{-3ex}
\caption{Overview of our proposed training-free CD-FSMIS model \textbf{MAUP}.}
\vspace{-3ex}
\label{the overview}
\end{figure*}

\section{Methodology}
\subsection{Overall Architecture}

The proposed MAUP strategy introduces a training-free approach for CD-FSMIS via leveraging the foundational model Segment Anything Model (SAM). Given the query image $\mathbf{I}_{q}$ and a selected support image $\mathbf{I}_s$ with corresponding support mask $\mathbf{M}_s$, our goal is to segment the target organs in $\mathbf{I}_{q}$ without any model training. The procedure begins with extracting discriminative features $\mathbf{F}_s$ and $\mathbf{F}_q$ from both query image $\mathbf{I}_q$ and support image $\mathbf{I}_s$ using the pre-trained DINOv2 encoder. These features are then employed to calculate similarity maps that guide the prompting strategy MAUP, which can dynamically create the optimal point prompts for SAM model.
As illustrated in Fig. \ref{the overview}, our method consists of three main steps: (1) feature extraction and similarity computation, (2) MAUP prompting strategy, and (3) final mask prediction.

\subsection{Feature Extraction and Similarity Map Computation}

We employ the frozen DINOv2 encoder \cite{oquab2023dinov2} which is pre-trained as feature extractor to obtain discriminative representations from both support and query images considering its robust and generalizable extraction capability. For each image, the features are first extracted through: $\mathbf{F}_s = \operatorname{E}(\mathbf{I}_s), \mathbf{F}_q = \operatorname{E}(\mathbf{I}_q)$, where $\operatorname{E}(\cdot)$ denotes the DINOv2 encoder. 

\begin{figure*}[!t]                   % htbp
\centering
\includegraphics[width=0.45\textwidth]{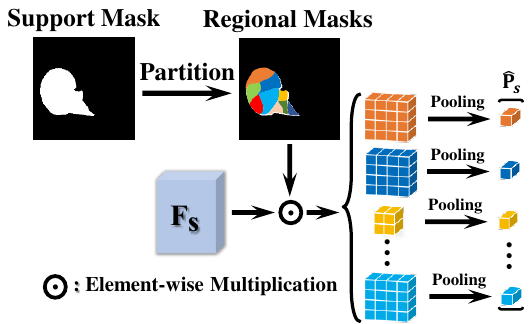}
\vspace{-3ex}
\caption{The Regional Prototype Generation (RPG) module.}
\vspace{-4ex}
\label{the RPG}
\end{figure*}

As shown in Fig. \ref{the RPG}, for subsequent multi-center positive prompting strategy, we employ the Region Prototype Generation (RPG) module used in RPT \cite{zhu2023few} to calculate multiple regional prototypes through the subdivisions in the foreground of the support image. Specifically, given a support image $\mathbf{I}_s$ and corresponding foreground mask $\mathbf{M}_s$, the foreground region of this image is obtained with calculating their production. Then, the Voronoi-based partition method \cite{aurenhammer1991voronoi} is employed to divide foreground into $N_f$ regions and forms a set of regional masks $\left \{\mathbf{V}_n \right \}^{N_f}_{n=1}$, where $N_f$ is set as $30$ according to the experiments. After that, a set of regional prototypes $\hat{\mathcal{P}}_{s} = \left \{  \hat{\mathbf{p}}_{n}\right \}^{N_f}_{n=1}, \hat{\mathbf{p}}_{n} \in \mathbb{R}^{1 \times C}$ are generated with use of Masked Average Pooling (MAP). Formally,        
\begin{equation}
\hat{\mathbf{p}}_{n} = \operatorname{MAP}(\mathbf{F}_s, \mathbf{V}_n) = \frac{1}{\left |  \mathbf{V}_n\right | } \sum_{i=1}^{HW}\mathbf{F}_{s,i} \mathbf{V}_{n,i},
\end{equation} 
where $\mathbf{F}_s \in  \mathbb{R}^{C \times H \times W}$ denotes the support feature extracted by DINOv2 encoder. With these regional prototypes, we can obtain the multi-center positive similarity maps set $\mathcal{S}_{p} = \left \{  S_{n}  \right \}^{N_f}_{n=1}$ through the cosine similarity computation:
\begin{equation}
\mathbf{S}_{n} = \operatorname{cos}(\mathbf{F}_q, \hat{\mathbf{p}}_{n}),
\end{equation} where $\mathbf{S}_{n} \in \mathbb{R}^{ H \times W}$ denotes the single similarity map of each region.

Besides, for the negative prompts, we propose to use the periphery area of organ to select effective negative prompts, which are able to delineate the boundaries of target structure and surrounding tissues. The periphery area is extracted through a morphological approach. To specific, we first apply the morphological dilation operation $\delta_{r}(\cdot)$ to the support mask $\mathbf{M}_s$ by using a circular structuring element with radius $r$, the dilated support mask $\mathbf{M}_{dilated}$ is calculated as:
\begin{equation}
    \mathbf{M}_{dilated} =  \delta_{r} (\mathbf{M}_s).
\end{equation}

Then, the periphery support mask $\tilde{\mathbf{M}}_{s}$ is obtained by subtracting the original support mask $\mathbf{M}_s$ from the dilated mask $\mathbf{M}_{dilated}$, written as: $\tilde{\mathbf{M}}_{s} = \mathbf{M}_{dilated} - \mathbf{M}_{s}$, where $\tilde{\mathbf{M}}_{s} \in \mathbb{R}^{ H \times W}$. Therefore, the periphery prototype is obtained by: $\tilde{\mathbf{P}} = \operatorname{MAP}(\mathbf{F}_s, \tilde{\mathbf{M}}_{s})$, where $\tilde{\mathbf{P}} \in \mathbb{R}^{1 \times C}$ and the negative similarity map is calculated by: $\tilde{\mathbf{S}} = \operatorname{cos} (\mathbf{F}_q, \tilde{\mathbf{P}})$.

\subsection{Multi-center Adaptive Uncertainty-aware Prompting}

\subsubsection{Positive Prompting Strategy.}
The positive prompting strategy of MAUP employs a two-path approach to identify the optimal point prompts that guide the SAM model to segment target regions. As illustrated in Fig. \ref{the overview}, we design a two-path of prompting strategy which consists of mean similarity map based prompting and uncertainty map based prompting. Specifically, the mean similarity map is calculated over similarity maps $\mathcal{S}_{p} = \left \{ \mathbf{S}_{n} \right \}_{n=1}^{N_f}$, written as:
\begin{equation}
    \mathbf{\upsilon}(x, y) = \frac{1}{N_{f}}\sum_{n=1}^{N_{f}} \mathbf{S}_{n} (x, y),
\end{equation} 
where $\mathbf{\upsilon} \in \mathbb{R}^{H \times W}$. Based on mean similarity map $\mathbf{\upsilon}$, we first perform pixel selection according to the highest similarity values from $\mathbf{\upsilon}$, which effectively identifies regions with the highest probability of containing the target anatomical structure. We employ $\mathcal{Q}_{mean}$ represent this set of candidate pixels: 
\begin{equation}
\mathcal{Q}_{mean} = \left \{ (x, y) | \mathbf{\upsilon}(x, y) \geq \tau_{mean} \right \},
\end{equation} 
where $\tau_{mean}$ is the threshold value corresponding to the 95th percentile of $\mathbf{\upsilon}$. To ensure spatial diversity among selected prompts, we apply K-means clustering to $\mathcal{Q}_{mean}$ with $k$ clusters. The centroids of these clusters are then selected as our mean similarity-based positive prompts, which ensures comprehensive coverage of the target region while avoiding redundant prompts in the same local area. The number of clusters $k$ is adaptively determined based on the complexity $C$ of target region, which is computed as: \begin{equation}
C =   \text{Area}(\mathbf{\upsilon}) +   \text{Perimeter}(\mathbf{\upsilon}),
\end{equation} 
where $\text{Area}(\cdot)$ and $\text{Perimeter}(\cdot)$ measure the area and boundary length of regions. Then, the number of clusters $k$ is determined as:
\begin{equation}
k = \text{max}(N_{min}, \text{min}(N_{max}, \lfloor \gamma \cdot C \rfloor)),
\end{equation} where $N_{min} = 3$ and $N_{max}=10$ are the minimum and maximum numbers of allowed prompts and $\gamma$ is a scaling factor. 

For the uncertainty map path, the uncertainty map is from the variance calculation over similarity maps $\mathcal{S}_{p} = \left \{ \mathbf{S}_{n} \right \}_{n=1}^{N_f}$ and identifies the challenging regions which require special attention, and the uncertainty map is calculated as: 
\begin{equation}
     \mathbf{U}(x, y) = \frac{1}{N_{f}} \sum_{n=1}^{N_{f}} (\mathbf{S}_{n}(x, y) - \mathbf{\upsilon}(x, y))^{2}, 
\end{equation} 
where $\mathbf{U} \in \mathbb{R}^{H \times W}$. We similarly perform the pixels selection with the highest uncertainty values from $\mathbf{U}$ representing regions where the model has the highest decision variance:
\begin{equation}
\mathcal{Q}_{uncert} = \left \{ (x, y) |\mathbf{U}(x, y) \geq \tau_{uncert} \right \},
\end{equation} where $\tau_{uncert}$ is the threshold value corresponding to the 95th percentile of $\mathbf{U}$. From $\mathcal{Q}_{uncert}$, we randomly select 2 points as uncertainty-based positive prompts. These prompts specifically target challenging regions that require special attention during segmentation. 

Finally, the positive point prompts set $\mathcal{Q}_{pos}$ can be obtained through merging two point prompt sets $\mathcal{Q}_{mean}$ and $\mathcal{Q}_{uncert}$ directly:
\begin{equation}
    \mathcal{Q}_{pos} = \mathcal{Q}_{mean} \cup \mathcal{Q}_{uncert}.
\end{equation}

\subsubsection{Negative Prompting Strategy.}
The negative prompting strategy is crucial for accurate boundary delineation, especially in medical images where target organs often have similar intensity profiles to surrounding tissues. As shown in Fig. \ref{the overview}, our negative prompting strategy leverages the periphery similarity map $\tilde{\mathbf{S}}$ to identify regions that should be excluded from the segmentation.

From the periphery similarity map $\tilde{\mathbf{S}}$, we perform the pixel selection according to the highest similarity values. These pixels represent regions in the query image that are most similar to the periphery of the target organ in the support image. Formally, the set of candidate negative prompts $\mathcal{Q}_{neg}$ is defined as:
\begin{equation}
\mathcal{Q}_{neg} = \left \{ (x, y) | \tilde{\mathbf{S}}(x, y) \geq \tau_{neg}  \right \},
\end{equation} where $\tau_{neg}$ is the threshold value corresponding to the 95th percentile of $\tilde{\mathbf{S}}$.

\subsection{Query Mask Prediction}
After generating both positive and negative prompts through  MAUP, the final step is to feed these prompts into the SAM model to obtain the segmentation mask for the query image. As illustrated in Fig. \ref{the overview}, the positive prompts guide the SAM model to include the target regions in the segmentation, while the negative prompts help exclude surrounding tissues and refine the boundaries.

\begin{table}[!t]
\centering
\caption{Quantitative Comparison (in Dice score $\%$) of different methods on \textbf{Abd-MRI} and \textbf{Abd-CT}. The best is shown in \textbf{bold} and the second-best is \underline{underlined}.}
\vspace{-1ex}
\label{Quantitative Results 1}
\resizebox{\textwidth}{!}{
\setlength{\tabcolsep}{3pt}
\begin{tabular}{l|l|ccccc|ccccc}
\toprule[1pt]
\midrule
\multirow{2}{*}{Method} & \multirow{2}{*}{Ref.} & \multicolumn{5}{c|}{\textbf{Abd-MRI}}           & \multicolumn{5}{c}{\textbf{Abd-CT}}             \\ \cmidrule{3-12} 
                        &                       & Liver & LK    & RK    & Spleen & Mean  & Liver & LK    & RK    & Spleen & Mean  \\ \midrule
PANet \cite{wang2019panet}   & ICCV'19               & 39.24 & 26.47 & 37.35 & 26.79  & 32.46 & 40.29 & 30.61 & 26.66 & 30.21  & 31.94 \\
SSL-ALP \cite{ssl-alp}       & TMI'22                & 70.74 & 55.49 & 67.43 & \underline{58.39}  & 63.01 & 71.38 & 34.48 & 32.32 & 51.67  & 47.46 \\
RPT \cite{zhu2023few}        & MICCAI'23             & 49.22 & 42.45 & 47.14 & 48.84  & 46.91 & 65.87 & 40.07 & 35.97 & 51.22  & 48.28 \\
PATNet \cite{lei2022cross}   & ECCV'22               & 57.01 & 50.23 & 53.01 & 51.63  & 52.97 & 75.94 & 46.62 & 42.68 & \underline{63.94}  & 57.29 \\
IFA \cite{nie2024cross}      & CVPR'24               & 50.22 & 35.99 & 34.00 & 42.21  & 40.61 & 46.62 & 25.13 & 26.56 & 24.85  & 30.79 \\
FAMNet \cite{bo2025cross}    & AAAI'25               & \underline{73.01} & \underline{57.28} & \textbf{74.68} & 58.21  & \underline{65.79} & \underline{73.57} & \underline{57.79} & \underline{61.89} & \textbf{65.78}  & \underline{64.75} \\
\rowcolor{gray!45} Ours       &           & \textbf{78.16} & \textbf{58.23} & \underline{72.34} & \textbf{59.65} & \textbf{67.09} & \textbf{78.25} & \textbf{59.41} & \textbf{71.80} & 60.38  & \textbf{67.46} \\ \midrule
\bottomrule[1pt]
\end{tabular}}
\vspace{-2ex}
\end{table}

\section{Experiments}

\subsection{Datasets and Implementation Details}
          
\textbf{Datasets.} We evaluate our proposed method on three diverse medical imaging datasets that represent different modalities and anatomical structures. To be specific, \textbf{Abd-MRI} consists of 20 cases of abdominal MRI scans from the ISBI 2019 Combined Healthy Organ Segmentation challenge (CHAOS) \cite{kavur2021chaos} and \textbf{Abd-CT} consists of 20 cases of abdominal CT scans from the MICCAI 2015 multi-atlas labeling Beyond The Cranial Vault challenge (BTCV) \cite{ABD-CT}. The \textbf{Card-MRI} contains 45 cases of cardiac MRI scans collected from the MICCAI 2019 Multi-Sequence Cardiac MRI Segmentation Challenge \cite{zhuang2018multivariate}. 

\begin{table}[!t]
\centering
\caption{Quantitative Comparison (in Dice score $\%$) of different methods on \textbf{Card-MRI}. The best value is shown in \textbf{bold} font, and the second-best value is \underline{underlined}.}
\vspace{-1ex}
\label{Quantitative Results 2}
\resizebox{0.7\textwidth}{!}{
\setlength{\tabcolsep}{6pt}
\begin{tabular}{l|l|cccc}
\toprule[1pt]
\midrule
\multirow{2}{*}{Method} & \multirow{2}{*}{Ref.} & \multicolumn{4}{c}{Card-MRI}  \\ \cmidrule{3-6} 
                        &                       & LV-BP & LV-MYO & RV    & Mean  \\ \midrule
PANet \cite{wang2019panet}           & ICCV'19               & 51.42 & 25.75  & 25.75 & 36.66 \\
SSL-ALP \cite{ssl-alp}               & TMI'22                & 83.47 & 22.73  & 66.21 & 57.47 \\
RPT \cite{zhu2023few}                & MICCAI'23             & 60.84 & 42.28  & 57.30 & 53.47 \\
PATNet \cite{lei2022cross}           & ECCV'22               & 65.35 & 50.63  & 68.34 & 61.44 \\
IFA \cite{nie2024cross}              & CVPR'24               & 50.43 & 31.32  & 30.74 & 37.50 \\
FAMNet \cite{bo2025cross}            & AAAI'25               & \underline{86.64} &\underline{51.82} & \underline{76.26} & \underline{71.58} \\
\rowcolor{gray!45} Ours           &                          & \textbf{88.36} & \textbf{52.74}   & \textbf{78.29} & \textbf{73.13}\\ \midrule
\bottomrule[1pt]
\end{tabular}}
\vspace{2ex}
\end{table}

\begin{figure*}[!t]                   % htbp
\centering
\includegraphics[width=0.99\textwidth]{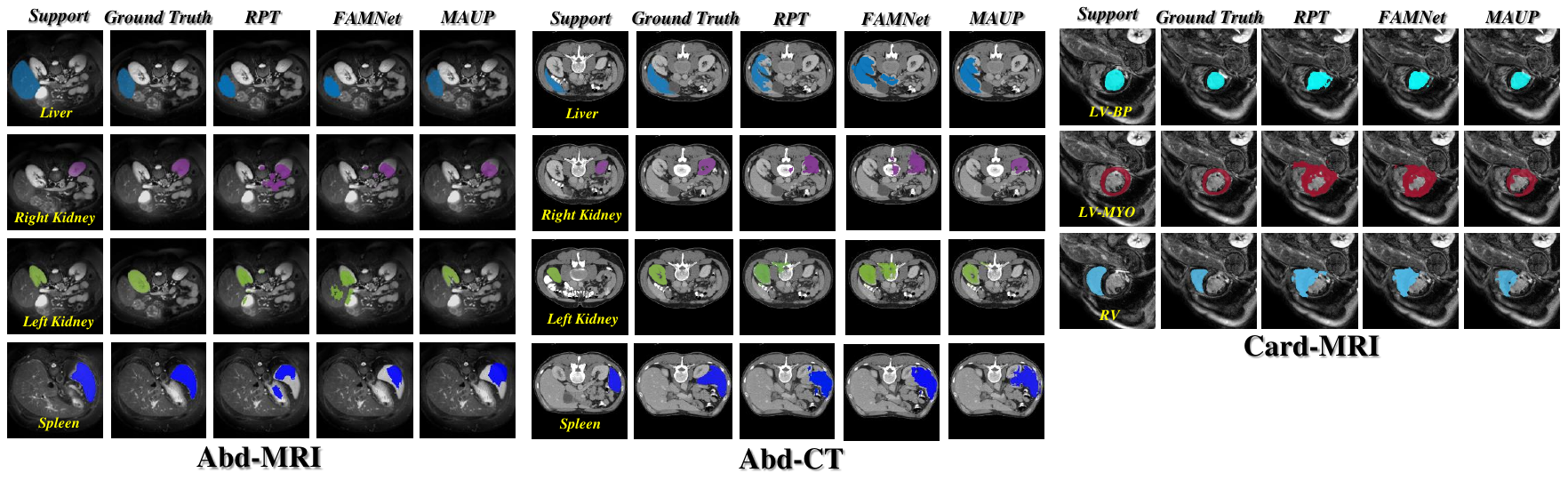}
\vspace{-2ex}
\caption{The qualitative results of our model and compared models in datasets: Abd-MRI, Abd-CT and Card-MRI.}
\vspace{-2ex}
\label{Qualitative results}
\end{figure*}

\noindent \textbf{Implementation Details.}
We employ the pretrained DINOv2-ViT-L/14 \cite{oquab2023dinov2} as the feature extractor $\operatorname{E}(\cdot)$ for the high-quality visual representation. For segmentation component, we utilize the SAM-ViT-H as encoder of the SAM model, which leads to the excellent segmentation performance in our task. For a comparison, the metric used to evaluate the performance of 2D slices on 3D volumetric ground-truth is the Dice score \cite{gmrd}. Our model is implemented with the PyTorch v1.12 framework, and all of experiments are conducted on a workstation with an NVIDIA 3090 GPU with 24GB memory. To simulate the scarcity of labeled data in medical scenarios, all experiments embrace the 1-way 1-shot setting.

\subsection{Quantitative and Qualitative Results}

Table \ref{Quantitative Results 1} and Table \ref{Quantitative Results 2} presents a comprehensive performance comparison between our proposed model and several Few-shot Medical Image Segmentation (FSMIS) models, including PANet \cite{wang2019panet}, SSL-ALP \cite{ouyang2020self}, RPT \cite{zhu2023few}, IFA \cite{nie2024cross} and FAMNet \cite{bo2025cross}. The average dice scores of target organs or regions are compared in this table. For a fair comparison, it is important to note that PANet, SSL-ALP, RPT, IFA and FAMNet all employ medical image datasets as source domain and fine-tune the models on them, while ours directly inferences on three target medical image datasets (Abd-MRI, Abd-CT and Card-MRI) in a training-free manner, which is much harder than these methods. From the two tables, it can be seen that the proposed method outperforms all listed methods in terms of the mean Dice values obtained on three datasets. Specifically, our model outperforms second-best models by $1.3 \%$, $2.71 \%$ and $1.55 \%$ on Abd-MRI, Abd-CT and Card-MRI. In addition to the quantitative comparisons, Fig. \ref{Qualitative results} shows qualitative results of our model compared to the alternative model on Abd-MRI, Abd-CT and Card-MRI datasets. The visual evidence clearly demonstrates boundary preservation capabilities and enhanced generalization performance of our model across different anatomical structures and imaging modalities.

\subsection{Ablation Studies}

The ablation studies are conducted on the Abd-MRI dataset. As illustrated in Fig. \ref{ablation of N_f}, the number of similarity maps $N_f$ (number of divided foreground regions) significantly influences model performance, with an optimal value yielding the highest Dice score. Table \ref{ablation of prompting} reveals notable performance differences when comparing Uncertainty Map based Prompts (UMP), Mean similarity Map based Prompts (MMP) and Negative Prompts (NP). Specifically, the combination of three prompting strategies achieves the highest Dice score compared to partial prompting strategies approach.

\begin{table}[!t]
\begin{minipage}{0.4\textwidth}
		\centering
  \vspace{-3ex}
        \caption{Ablation study with different prompting strategies on \textbf{Abd-MRI}.}
        \label{ablation of prompting}
        \begin{tabular}{ccc|c}
         \toprule[1pt]
          \midrule
            UMP & MMP & NP & Dice Score ($\%$) \\ \midrule
           \checkmark   &     &             & 65.08   \\
            \checkmark   & \checkmark   &    & 66.71   \\
            \rowcolor{gray!45} \checkmark   & \checkmark   & \checkmark  & \textbf{67.09}   \\ \midrule
             \bottomrule[1pt]
         \end{tabular}
	\end{minipage} \hfill \quad
    \begin{minipage}{0.57\textwidth}
		\centering
		\includegraphics[width=0.8\textwidth]{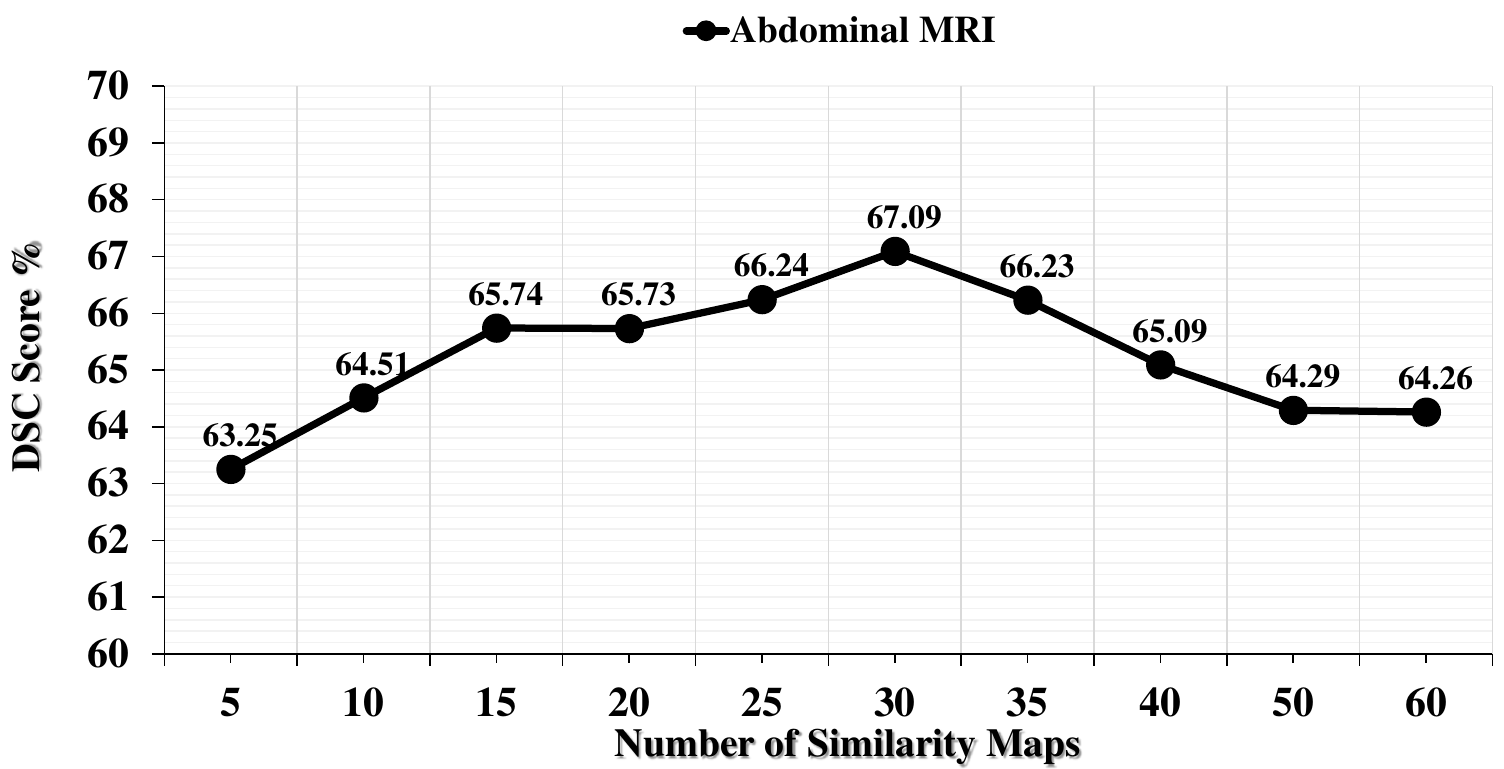}	\makeatletter\def\@captype{figure}\makeatother\caption{Ablation study the of number of similarity maps $N_f$.}
		\label{ablation of N_f}		
	\end{minipage} 
 \vspace{-4ex}
\end{table}

% \begin{figure*}[!ht]                   % htbp
% \centering
% \includegraphics[width=0.6\textwidth]{ablation.pdf}
% \vspace{-2ex}
% \caption{The ablation study of number of similarity maps $N_f$.}
% \vspace{-1ex}
% \label{ablation of N_f}
% \end{figure*}

% \begin{table}[]
% \centering
% \caption{Ablation study of prompting strategies.}
% \vspace{-1ex}
% \label{ablation of prompting}
% \begin{tabular}{ccc|c}
% \toprule[1pt]
% \midrule
% UMP & MMP & NP & Dice Score ($\%$) \\ \midrule
% \checkmark   &     &             & 62.08   \\
% \checkmark   & \checkmark   &    & 64.61   \\
% \rowcolor{gray!45} \checkmark   & \checkmark   & \checkmark  & \textbf{65.82}   \\ \midrule
% \bottomrule[1pt]
% \end{tabular}
% \vspace{-3ex}
% \end{table}

\section{Conclusion}

In this paper, we proposed MAUP, a training-free few-shot medical image segmentation approach that introduces a multi-center adaptive uncertainty-aware prompting strategy for the SAM model. By combining spatial diversity, uncertainty guidance, and adaptive prompt optimization, MAUP effectively addresses the challenges of medical image segmentation without requiring any additional training on medical datasets. Our comprehensive experiments on three diverse medical imaging datasets demonstrate that MAUP outperforms conventional training-based few-shot segmentation models under the training-free manner. The proposed method achieves precise segmentation results across different imaging modalities and anatomical structures, highlighting its potential for real-world clinical applications.

\begin{credits}
\subsubsection{\ackname} 
This work was supported by the National Natural Science Foundation of China under Grants 62371235.

\subsubsection{\discintname}
All authors declare that they have no conflicts of interest
\end{credits}

\bibliographystyle{splncs04}
\bibliography{references}

\end{document}